\documentclass[conference]{IEEEtran}

\usepackage{cite}
\usepackage{amsmath,amssymb,amsfonts,amsthm,mathtools}
\usepackage{booktabs}
\usepackage{graphicx}
\usepackage{textcomp}
\usepackage{xcolor}
\usepackage{multirow}
\usepackage{adjustbox}
\usepackage{url}
\usepackage{balance}
\usepackage{hyperref}
\usepackage[linesnumbered,ruled,vlined]{algorithm2e}
\usepackage[compatibility=false]{caption}
\usepackage{subcaption}

\newtheorem{lemma}{Lemma}

\DeclareMathOperator{\nnz}{nnz}

\definecolor{bestgreen}{RGB}{0,150,0}
\definecolor{secondblue}{RGB}{0,0,200}
\newcommand{\best}[1]{\textbf{\textcolor{bestgreen}{#1}}}
\newcommand{\second}[1]{\textcolor{secondblue}{#1}}

\def\BibTeX{{\rm B\kern-.05em{\sc i\kern-.025em b}\kern-.08em
    T\kern-.1667em\lower.7ex\hbox{E}\kern-.125emX}}

\begin{document}

\title{Randomized SVD Approximations for Spectral Co-Clustering of Word-Document Matrices}

\author{
\IEEEauthorblockN{Fateme Mazdarani}
\IEEEauthorblockA{
School of Computing\\
Clemson University\\
Clemson, SC, USA\\
fmazdar@clemson.edu
}
\and
\IEEEauthorblockN{Carlos Toxtli}
\IEEEauthorblockA{
School of Computing\\
Clemson University\\
Clemson, SC, USA\\
ctoxtli@clemson.edu
}
}

\maketitle

\begin{abstract}
Spectral co-clustering is a useful tool for discovering latent structure in word-document matrices, but its reliance on singular value decomposition (SVD) can make standard formulations expensive on high-dimensional data. This paper presents two randomized approximations for normalized spectral co-clustering of bipartite text data when the numbers of document and word clusters may differ. The first method uses randomized SVD through random projection, while the second combines partial SVD with element-wise random sampling. Across real-world and synthetic datasets, both methods reduce runtime relative to the full-SVD baseline, but their behavior depends on matrix sparsity. The random-projection method is the more reliable approximation across the tested settings, whereas the sampling-based method is most useful on denser matrices and provides limited benefit on already sparse text data. These results show that randomized approximations for spectral co-clustering should be selected according to the underlying structure of the data.
\end{abstract}


\section{Introduction}

Modern data mining and natural language processing applications involve increasingly large and high-dimensional text collections. Clustering is widely used to uncover latent structure by grouping similar data points. Document clustering has applications in organization, retrieval, and exploratory analysis \cite{aggarwal2012}. However, standard single-view clustering methods such as $k$-means, fuzzy $c$-means, and expectation-maximization cluster documents independent of the vocabulary structure. Spectral clustering \cite{luxburg2007} addresses some limitations of centroid-based methods by embedding data with eigenvectors of a graph Laplacian. It has been widely used for document clustering because it can capture non-convex structure in similarity graphs. However, eigen decomposition can be expensive in both time and memory \cite{chen2011}. This motivates scalable spectral approximations for text data. A term-document matrix naturally represents two coupled entity types: documents and words. Clustering only one side of the matrix ignores these dependencies. Joint row-column clustering is therefore useful when the goal is to identify groups of documents together with the vocabulary patterns that characterize them.

Spectral co-clustering models the term-document matrix as a bipartite graph and clusters documents and words simultaneously \cite{dhillon2001}. The method computes singular vectors of a normalized matrix and applies $k$-means to the resulting document and word embeddings. This captures relations between document groups and word groups rather than treating them as independent clustering problems. The original bipartite spectral formulation uses a shared partitioning structure, whereas many practical term-document matrices may require different numbers of document and word clusters. Related spectral biclustering methods, especially in bioinformatics, also seek simultaneous row and column partitions and can model checkerboard patterns with different numbers of row and column clusters \cite{cheng2000,kluger2003}. These methods are useful for structured matrices, but their reliance on singular-vector computations limits direct use on high-dimensional text matrices.

The main computational bottleneck is the SVD step. For an $n\times m$ matrix, exact SVD can be costly when both dimensions are large. Randomized SVD reduces this cost by using random projections to approximate the dominant singular subspaces \cite{halko2011}. Partial SVD methods compute only leading singular components and are especially relevant for sparse term-document matrices
\cite{manning2008}.

Motivated by these observations, we study randomized and sparsity-aware approximations within a normalized spectral co-clustering pipeline for word-document matrices. The goal is not to introduce a new co-clustering objective, but to examine how different SVD approximations affect runtime and clustering quality when row and column cluster counts may differ. The main contributions are: (i) a normalized bipartite spectral co-clustering pipeline for nonsquare word-document matrices with different document and word cluster counts; and (ii) an empirical comparison of random-projection SVD and partial-SVD sampling variants against a full-SVD baseline on real and synthetic matrices.

\section{Related Work}

Co-clustering jointly partitions rows and columns of a data matrix, with classical approaches including direct matrix co-clustering, spectral biclustering, bipartite spectral graph partitioning, information-theoretic co-clustering, Bregman co-clustering, and matrix-factorization methods \cite{hartigan1972,cheng2000,kluger2003,dhillon2001,dhillon2003_itcc,banerjee2007_bregman,xu2003,xu2004}.
Scalability remains a major challenge \cite{wang2024_survey_coclustering}. Our work keeps the normalized spectral formulation and studies how SVD approximations affect runtime and clustering quality.

Graph-based co-clustering models two entity types as a bipartite or heterogeneous graph. Related methods include document-word spectral graph models \cite{zha2001}, semi-supervised spectral co-clustering \cite{shi2010}, structured optimal bipartite graph learning \cite{nie2017}, and fast flexible bipartite graph partitioning with different numbers of row and column clusters \cite{chen2023}. Recent scalable methods use anchor graphs or normalized bipartite cuts to reduce graph construction or eigensolver costs \cite{xie2025_fcals,nie2025_fdc2}. These works usually modify the graph model, objective, or optimization procedure. In contrast, we retain the standard normalized spectral co-clustering pipeline and evaluate approximation choices within it.

Randomized and sparse spectral approximations are closely related to our method. Randomized SVD approximates dominant singular subspaces using random projections, oversampling, and power iterations \cite{halko2011}, and randomized spectral methods have been studied for large undirected networks and directed-network co-clustering \cite{zhang2022,guo2023}. Sparse SVD methods are important for text data because term-document matrices are high-dimensional and sparse; truncated SVD has long been used in latent semantic analysis \cite{manning2008}, and iterative methods such as Lanczos-type algorithms and IRLBA compute leading singular vectors with cost tied to the number of nonzero entries \cite{baglama2005}. Matrix sampling and sparsification have also been studied for randomized low-rank approximation \cite{frieze2004,drineas2006}. We combine these approximation tools with normalized spectral co-clustering for nonsquare word-document matrices and examine when sampling helps or removes useful signal from already sparse text data.

\section{Problem Statement}

\subsection{Word-Document Graph Modeling}

Following prior work \cite{dhillon2001,zha2001}, we model a corpus as an undirected bipartite graph $G=(D\cup W,E)$, where $D=\{d_1,\ldots,d_n\}$ is the set of documents and $W=\{w_1,\ldots,w_m\}$ is the vocabulary. An edge $(d_i,w_j)\in E$ indicates that word $w_j$ appears in document $d_i$; no edges connect two documents or two words. Edge weights encode word-document relevance. We use TF-IDF \cite{salton1983}, where term frequency measures local word frequency and inverse document frequency down-weights globally common words:
\begin{equation}
\mathrm{TF\mbox{-}IDF}(w_j,d_i)=\mathrm{TF}(w_j,d_i)\times\mathrm{IDF}(w_j),
\end{equation}
where $\mathrm{TF}(w_j,d_i)$ is the frequency of $w_j$ in $d_i$, and
\begin{equation}
\mathrm{IDF}(w_j)=\log\left(\frac{n}{1+n_j}\right),
\end{equation}
with $n$ denoting the number of documents and $n_j$ the number of documents containing $w_j$. The weighted adjacency matrix $A\in\mathbb{R}^{n\times m}$
is then
\begin{equation}
A_{ij}=\begin{cases}
E_{ij}, & \text{if word $w_j$ appears in document $d_i$},\\
0, & \text{otherwise},
\end{cases}
\end{equation}
where $E_{ij}$ is the corresponding TF-IDF weight.

\subsection{Graph Partitioning Problem}

Co-clustering can be viewed as partitioning the bipartite vertex set $\mathcal{V}=D\cup W$. For two subsets, the graph cut is
\begin{equation}
\min_{\mathcal{V}_1,\mathcal{V}_2}\mathrm{cut}(\mathcal{V}_1,\mathcal{V}_2)
=
\min_{\mathcal{V}_1,\mathcal{V}_2}
\sum_{v_i\in\mathcal{V}_1,v_j\in\mathcal{V}_2}w_{ij}.
\end{equation}
For a $k$-way partition, this becomes
\begin{equation}
\mathrm{cut}(\mathcal{V}_1,\ldots,\mathcal{V}_k)
=
\sum_{i<j}\mathrm{cut}(\mathcal{V}_i,\mathcal{V}_j).
\end{equation}
Because unconstrained cut minimization can produce degenerate unbalanced partitions, we use the normalized cut objective \cite{shi2000}:
\begin{equation}
\mathrm{Ncut}(\mathcal{V}_1,\ldots,\mathcal{V}_k)
=
\sum_{i=1}^{k}
\frac{\mathrm{cut}(\mathcal{V}_i,\mathcal{V}-\mathcal{V}_i)}
{\mathrm{weight}(\mathcal{V}_i)},
\end{equation}
where $\mathrm{weight}(\mathcal{V}_i)=\sum_{v\in\mathcal{V}_i}\sum_{u\in\mathcal{V}}w_{vu}$. Minimizing Ncut partitions the bipartite graph into paired document-word groups $\mathcal{V}_i=D_i\cup W_i$. Since exact graph partitioning is NP-complete, spectral co-clustering uses a continuous relaxation:
\begin{equation}
\min_S\operatorname{Tr}(S^\top LS)\quad\mathrm{s.t.}\quad S^\top S=I_k.
\end{equation}
By the Rayleigh-Ritz theorem, this relaxation is solved by eigenvectors associated with the smallest eigenvalues of $L$. For efficiency, we work with the normalized adjacency matrix. Let $A\in\mathbb{R}^{n\times m}$ be the weighted bipartite adjacency matrix. Define $A_n=D_r^{-1/2}AD_c^{-1/2}$, where $D_r\in\mathbb{R}^{n\times n}$ and $D_c\in\mathbb{R}^{m\times m}$ are row and column degree matrices: $ (D_r)_{ii}=\sum_{j=1}^{m}A_{ij}, (D_c)_{jj}=\sum_{i=1}^{n}A_{ij} $. The following standard relationship connects the normalized-cut relaxation to the SVD of $A_n$.

\begin{lemma}
Let $A\in\mathbb{R}^{n\times m}$, with degree matrices $D_r\in\mathbb{R}^{n\times n}$ and $D_c\in\mathbb{R}^{m\times m}$. Define $L=\begin{bmatrix}D_r&-A\\-A^\top&D_c\end{bmatrix}$, $D=\begin{bmatrix}D_r&0\\0&D_c\end{bmatrix}$, and $A_n=D_r^{-1/2}AD_c^{-1/2}$. If $Lz=\lambda Dz$ with $z=\begin{bmatrix}x\\y\end{bmatrix}$, then $u=D_r^{1/2}x$ and $v=D_c^{1/2}y$ satisfy $A_nv=(1-\lambda)u$ and $A_n^\top u=(1-\lambda)v$. Thus $\sigma=1-\lambda$ is a singular value of $A_n$, and $x=D_r^{-1/2}u$, $y=D_c^{-1/2}v$.
\end{lemma}

\section{Method}

\subsection{Baseline Normalized Spectral Co-clustering}

We use normalized spectral co-clustering as the baseline framework and compare two SVD approximations with it. Given $A\in\mathbb{R}^{n\times m}$, normalized spectral co-clustering first forms $A_n=D_r^{-1/2}AD_c^{-1/2}$, computes leading singular vectors of $A_n$, and then clusters the document and word embeddings separately. Let $l_r=\lceil\log_2 k_r\rceil$ and $l_c=\lceil\log_2 k_c\rceil$. Following the multi-way construction of Dhillon~\cite{dhillon2001}, we skip the trivial first singular vector and use $U(:,2:l_r+1)$ for documents and $V(:,2:l_c+1)$ for words. Algorithm~\ref{nscc} is our full-SVD baseline. Compared with the original equal-partition setting \cite{dhillon2001}, we allow $k_r\neq k_c$ by using separate embedding dimensions $l_r$ and $l_c$ and running $k$-means independently on the document and word embeddings.

\begin{algorithm}[t]
\caption{Normalized Spectral Co-Clustering}
\label{nscc}
\KwIn{$A\in\mathbb{R}^{n\times m}$, row clusters $k_r$, column clusters $k_c$}
\KwOut{predicted row and column labels: \texttt{row\_labels},\texttt{col\_labels}}
Compute diagonal matrices $D_r,D_c$ such that $D_r(i,i)=\sum_j A(i,j)$, $D_c(i,i)=\sum_j A(j,i)$\;
Calculate the normalized matrix $A_n$ such that $A_n=D_r^{-1/2}AD_c^{-1/2}$\;
Compute SVD for $A_n=U\Sigma V^\top$\;
Set $l_r=\lceil\log_2 k_r\rceil$, $l_c=\lceil\log_2 k_c\rceil$\;
Set $X_r=D_r^{-1/2}U(:,2:l_r+1)$, $X_c=D_c^{-1/2}V(:,2:l_c+1)$\;
Find \texttt{row\_labels}, and \texttt{col\_labels} By running $k$-means on $X_r$ and $X_c$ with $k_r$ and $k_c$ clusters\;
\end{algorithm}

\subsection{Spectral Co-Clustering with Random Projection (SCCRP)}

SCCRP method reduces the computational cost of spectral co-clustering by approximating the singular vectors of the normalized matrix using randomized SVD. For a non-square matrix $A\in\mathbb{R}^{n\times m}$ and target rank $k$, the goal is to construct low-dimensional orthonormal bases $Q_1$ and $Q_2$ that capture the dominant column and row spaces of $A$, so that $A\approx Q_1Q_1^\top AQ_2Q_2^\top$. The matrices $Q_1$ and $Q_2$ are obtained by projecting $A$ and $A^\top$ onto random Gaussian test matrices, optionally using power iterations to improve accuracy when the singular values decay slowly. After orthonormalizing the projected matrices with QR decomposition, SCCRP computes the SVD of the smaller matrix $B=Q_1^\top AQ_2$. If $B=\widetilde U\Sigma\widetilde V^\top$, then the approximate singular vectors of $A$ are given by $U=Q_1\widetilde U$ and $V=Q_2\widetilde V$. This method also uses power iterations $(AA^\top)^qA\Omega_1$ and $(A^\top A)^qA^\top\Omega_2$, to increase the approximation accuracy, especially when the singular values of $A$ decay slowly. The randomized SVD procedure is given in Algorithm~\ref{rpsvd}, and its integration into the normalized spectral co-clustering pipeline is shown in Algorithm~\ref{sccrp}. The randomized projection step preserves the dominant singular subspaces with high probability when the target rank and number of power iterations are sufficiently large \cite{halko2011}. We use $q=2$ power iterations in our experiments.

If $\|A_n-\widehat{A}_n\|_2\leq\epsilon$ and the retained singular subspace has gap $\gamma>0$, Wedin-type perturbation bounds give a subspace error of order $O(\epsilon/\gamma)$. Thus, when the approximation error is small relative to the
singular-value gap and the embedded clusters are sufficiently separated, the resulting $k$-means assignments are expected to remain stable.

\begin{algorithm}[t]
\caption{Random Projection SVD}
\label{rpsvd}
\KwIn{$A\in\mathbb{R}^{n\times m}$: input matrix,\\
$k$: target number of singular vectors,\\
$q$: number of power iterations}
\KwOut{$U,\Sigma,V$ where $U\Sigma V^\top$ is a rank-$k$ approximation of $A$.}
Generate Gaussian test matrices $\Omega_1\sim\mathcal{N}(0,1)^{m\times2\times k}$ and $\Omega_2\sim\mathcal{N}(0,1)^{n\times2\times k}$\;
Form $Y_1=(AA^\top)^qA\Omega_1$ and $Y_2=(A^\top A)^qA^\top\Omega_2$ by multiplying alternately with $A$ and $A^\top$\;
Construct matrices $Q_1$ and $Q_2$ whose columns form orthonormal bases for the ranges of $Y_1$ and $Y_2$, respectively\;
Form the smaller matrix $B=Q_1^\top AQ_2$\;
Compute the SVD $B=\widetilde U \widetilde \Sigma\widetilde V^\top$\;
Retain $\widetilde{U}_k=\widetilde{U}(:,1:k)$, 
$\widetilde{V}_k=\widetilde{V}(:,1:k)$, and
$\Sigma=\widetilde{\Sigma}(1:k,1:k)$\;
Set $U=Q_1\widetilde U_k$ and $V=Q_2\widetilde V_k$\;
\end{algorithm}

\begin{algorithm}[t]
\caption{Spectral Co-Clustering with Random Projection-Based SVD}
\label{sccrp}
\KwIn{$A\in\mathbb{R}^{n\times m}$, row clusters $k_r$, column clusters $k_c$}
\KwOut{predicted row and column labels: \texttt{row\_labels},\texttt{col\_labels}}
Compute diagonal matrices $D_r,D_c$ such that $D_r(i,i)=\sum_j A(i,j)$, $D_c(i,i)=\sum_j A(j,i)$\;
Calculate the normalized matrix $A_n=D_r^{-1/2}AD_c^{-1/2}$\;
Set $l_r=\lceil\log_2 k_r\rceil$,
$l_c=\lceil\log_2 k_c\rceil$,
$k=\max(l_r,l_c)+1$\;
Approximate SVD for $A_n=U\Sigma V^T$ using random projection SVD method as described in Algorithm~\ref{rpsvd}\;
$U_k=D_r^{-1/2}U(:,2:l_r+1), V_k=D_c^{-1/2}V(:,2:l_c+1)$\;
Find \texttt{row\_labels}, and \texttt{col\_labels} By running $k$-means on $U_k$, and $V_k$, respectively\;
\end{algorithm}

The dominant cost of spectral co-clustering comes from computing the SVD, which SCCRP replaces with a randomized projection-based approximation. Its main cost is the sequence of matrix multiplications used in the power iterations, $(AA^\top)^qA\Omega_1$ and $(A^\top A)^qA^\top\Omega_2$. For sparse $A$, these multiplications require time of
$
O\left((2q+1)\lVert A\rVert_0(k+\max(m,n))\right)
$
The remaining SVD step is inexpensive because it is performed on a smaller projected matrix. Randomized SVD is also suited for parallelization \cite{halko2011}, making this approach practical for large-scale data.

\subsection{Spectral Co-Clustering with Partial SVD (SCCP)} 

\subsubsection{Partial SVD}
For sparse data, spectral co-clustering can avoid computing the full SVD by extracting only the top $k\ll\min(m,n)$ singular values and vectors: $A\approx U_k\Sigma_kV_k^\top$. Here, $U_k$ and $V_k$ contain the dominant left and right singular vectors, and $\Sigma_k$ contains the corresponding singular values. This approximation captures the main structure of the data while reducing noise and redundancy.

Partial SVD can be computed efficiently for sparse matrices using iterative methods such as Lanczos Bidiagonalization and the Implicitly Restarted Lanczos Bidiagonalization Algorithm (IRLBA) \cite{baglama2005}, which constructs a low-dimensional Krylov subspace and iteratively approximates dominant singular components while controlling memory usage. Its computational cost is approximately $O(k\cdot\nnz(A)\cdot t)$, where $\nnz(A)$ is the number of nonzero entries and $t$ is the number of iterations required for convergence. Since $k$ and $t$ are usually small, the runtime is nearly linear in the number of nonzero entries.

\subsubsection{Random Sampling}
As cost of partial SVD depends on $\nnz(A)$, we reduce it using element-wise random sampling. Unlike row or column sampling methods
\cite{frieze2004,drineas2006}, this strategy independently retains each nonzero entry with probability $p$, producing a uniformly sparse matrix
$A^{rs}$:
\[
A^{rs}_{ij}=\begin{cases}
A_{ij}, & \text{with probability }p,\\
0, & \text{with probability }1-p.
\end{cases}
\]

\begin{algorithm}[t]
\caption{Spectral Co-Clustering with Partial SVD and Random Sampling}
\label{sccpr}
\KwIn{$A\in\mathbb{R}^{n\times m}$, row clusters $k_r$, column clusters $k_c$, sampling probability $p$}
\KwOut{Predicted row and column labels: \texttt{row\_labels}, \texttt{col\_labels}}
Construct the sparsified matrix $A^{rs}$ using Equation (IV-C2)\;
Compute diagonal matrices $D_r$ and $D_c$ from the row and column sums of $A^{rs}$\;
Normalize the matrix as $A_n=D_r^{-1/2}A^{rs}D_c^{-1/2}$\;
Compute the partial SVD $A_n\approx U_k\Sigma_kV_k^\top$, where $k=\max(l_r,l_c)+1$, $l_r=\lceil\log_2 k_r\rceil$, $l_c=\lceil\log_2 k_c\rceil$\;
Set $U_k=U_k(:,2:l_r+1)$ and $V_k=V_k(:,2:l_c+1)$\;
Run $k$-means on $D_r^{-1/2}U_k$ and $D_c^{-1/2}V_k$ to obtain \texttt{row\_labels} and \texttt{col\_labels}, respectively\;
\end{algorithm}

\noindent Although this unscaled sampler satisfies $\mathbb{E}[A^{rs}]=pA$, the following degree normalization cancels any uniform multiplicative shrinkage: $(pD_r)^{-1/2}(pA)(pD_c)^{-1/2}=D_r^{-1/2}AD_c^{-1/2}$. Thus, the missing $1/p$ factor does not affect the normalized spectral structure under uniform shrinkage. When $p=1$, the method reduces to standard spectral co-clustering with partial SVD. Smaller values of $p$ increase sparsity and reduce runtime, while sufficiently large $p$ can preserve enough information for accurate clustering, consistent with prior work on randomized low-rank approximation and matrix sparsification~\cite{frieze2004,drineas2006}.

\subsection{Datasets and Metrics}

\subsubsection{Real-world data}
We use the 20 Newsgroups dataset~\cite{lang1995} containing documents from 20 categories. Since the dataset provides document labels but not word labels, we construct proxy word labels by clustering pretrained GloVe word embeddings. Proxy measures are commonly used for cluster validation
\cite{shayan2025functional}. We then build four TF-IDF word-document matrices with different document/word cluster counts, shown in Table~\ref{dataset}.

\begin{table}[tb]
\centering
\caption{Details of the dataset collections}
\label{dataset}
\scriptsize
\begin{tabular}{lccc}
\toprule
\textbf{Dataset} & \textbf{\#docs $\times$ \#words} & \textbf{Doc Clusters} & \textbf{Word Clusters}\\
\midrule
Newsgroups $3\times5$ & $2811\times198$ & 3 & 5\\
Newsgroups $4\times5$ & $3840\times766$ & 4 & 5\\
Newsgroups $3\times2$ & $2884\times4010$ & 3 & 2\\
Newsgroups $4\times3$ & $3843\times4158$ & 4 & 3\\
\bottomrule
\end{tabular}
\vspace{1ex}

{\raggedright Each subset was independently vectorized and filtered, resulting in different retained vocabulary sizes and matrix shapes. \par}
\end{table}

\begin{figure}[t]
\centering
\begin{subfigure}[b]{0.48\linewidth}
\centering
\includegraphics[width=\linewidth]{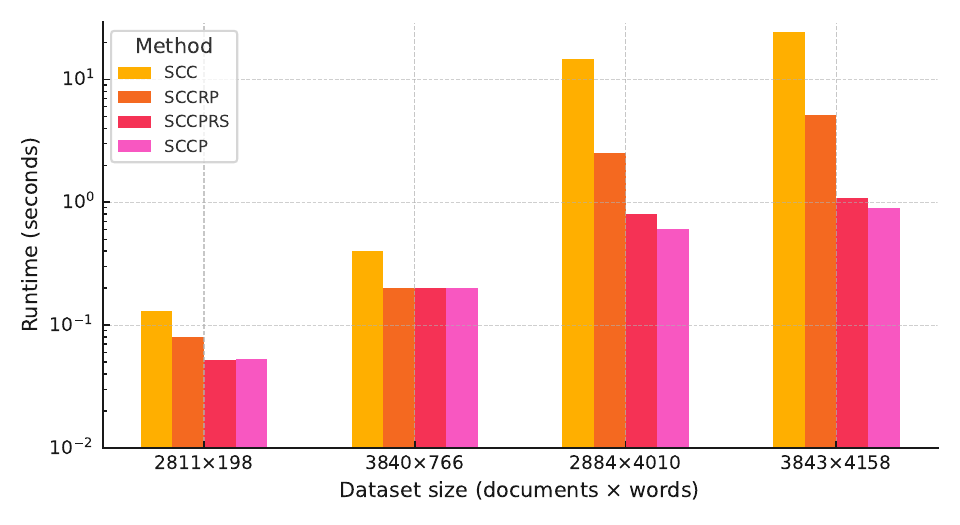}
\caption{Newsgroups}
\label{newsgroup_runtime}
\end{subfigure}
\hfill
\begin{subfigure}[b]{0.48\linewidth}
\centering
\includegraphics[width=\linewidth]{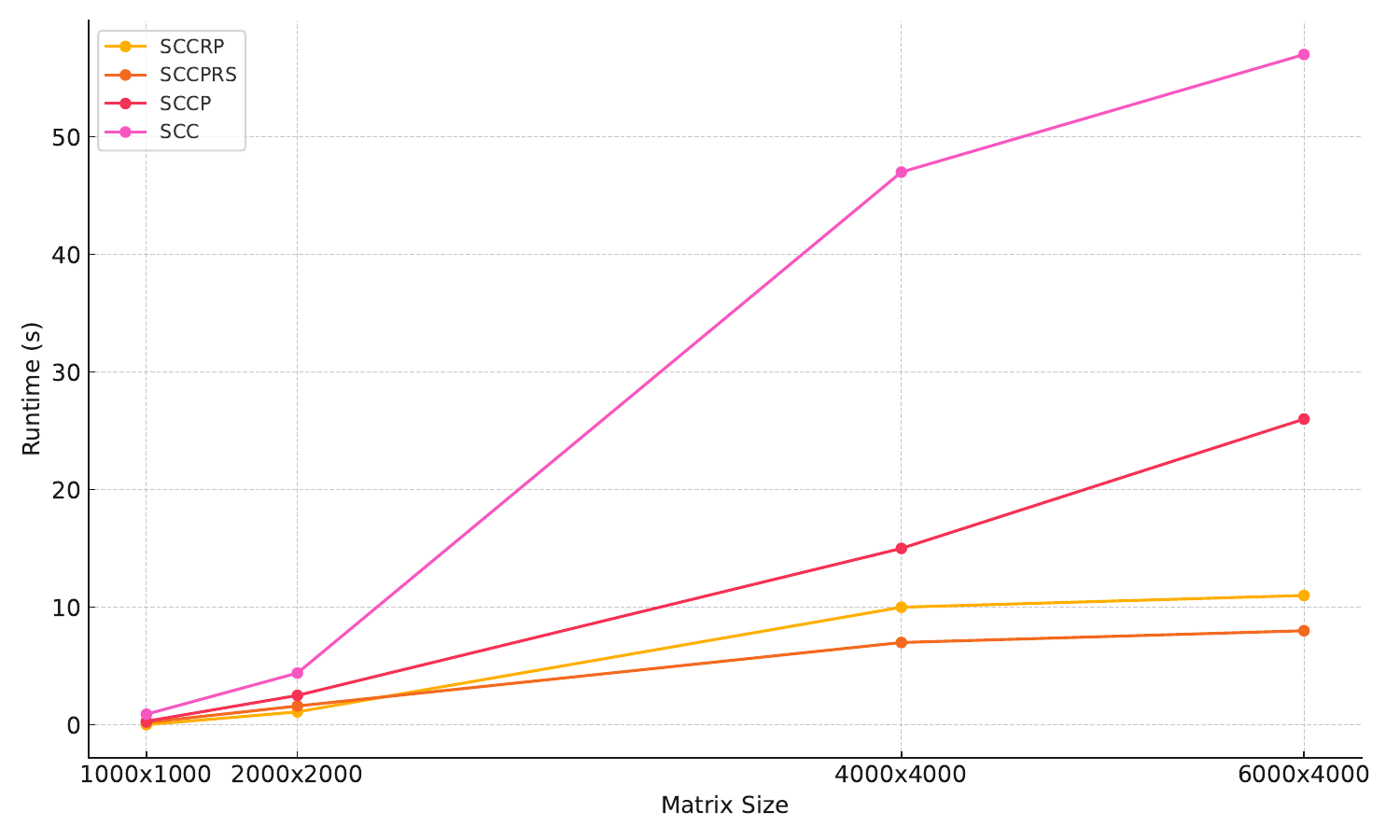}
\caption{Synthetic checkerboard}
\label{synthetic_runtime}
\end{subfigure}
\caption{Runtime comparison of spectral co-clustering variants. (a) Log-scale runtime on Newsgroups-derived matrices, where SCCRP and SCCP reduce runtime compared with SCC, while SCCP-RS provides little additional benefit. (b) Runtime on synthetic Checkerboard4x3 matrices of increasing size, where randomized methods show improved scalability.}
\end{figure}

\begin{table}[b]
\centering
\caption{Runtime (in seconds) of spectral co-clustering variants across Newsgroups datasets.}
\label{runtime_newsgroup}
\scriptsize
\begin{tabular}{lccccc}
\toprule
\textbf{Dataset} & \textbf{Size} & \textbf{SCC} & \textbf{SCCRP} & \textbf{SCCP-RS} & \textbf{SCCP}\\
\midrule
Newsgroups $3\times5$ & $2811\times198$ & 0.13 & 0.08 & 0.052 & 0.053\\
Newsgroups $4\times5$ & $3840\times766$ & 0.4 & 0.2 & 0.2 & 0.2\\
Newsgroups $3\times2$ & $2884\times4010$ & 14.6 & 2.5 & 0.8 & 0.6\\
Newsgroups $4\times3$ & $3843\times4158$ & 24.2 & 5.1 & 1.08 & 0.9\\
\bottomrule
\end{tabular}
\end{table}

\subsubsection{Synthetic data}
We also use checkerboard matrices generated by \texttt{make\_checkerboard} in \texttt{scikit-learn} with added Gaussian noise to simulate real-world imperfections. These matrices provide planted row and column clusters and allow controlled runtime comparisons across matrix sizes.

\subsubsection{Metrics}
We report row/column clustering accuracy (RAcc/CAcc), adjusted Rand index (R-ARI/C-ARI) \cite{vinh2010}, and wall-clock runtime. 
Accuracy is computed after exact matching between predicted and
reference cluster labels by enumerating all label permutations. Because the
evaluated cluster counts are at most five, exhaustive permutation matching is
tractable and returns the optimal assignment. ARI corrects pairwise clustering
agreement for chance.
All row and column metrics are computed separately. To account for randomness from randomized projection, element-wise sampling, and $k$-means initialization, each experiment is repeated over 20 random seeds, and all reported results are averaged across runs.

\subsubsection{Compared methods}
We compare SCC (full-SVD baseline, Algorithm~\ref{nscc}), SCCRP (randomized SVD, Algorithm~\ref{sccrp}), SCCP (partial SVD with $p=1$), and SCCP-RS (partial SVD with element-wise random sampling, Algorithm~\ref{sccpr}).

All methods were implemented in MATLAB and evaluated on the same local desktop PC using MATLAB's built-in SVD, sparse-matrix, and $k$-means routines with its default BLAS/LAPACK backend. Source code and settings are available at github.com/mazdarani/randomized-spectral-co-clustering.

\begin{table*}[t]
\centering
\caption{Clustering performance on Newsgroups-derived matrices. Each dataset block reports RAcc, CAcc, R-ARI, and C-ARI as an average over 20 random seeds. The best result for each metric is in \textcolor{bestgreen}{green}, and the second-best in \textcolor{secondblue}{blue}.}
\label{clustering_performance}
\scriptsize
\begin{tabular}{lcccccccccccccccc}
\toprule
\multirow{2}{*}{\textbf{Method}}
& \multicolumn{4}{c}{\textbf{Newsgroups 3x5}}
& \multicolumn{4}{c}{\textbf{Newsgroups 4x3}}
& \multicolumn{4}{c}{\textbf{Newsgroups 3x2}}
& \multicolumn{4}{c}{\textbf{Newsgroups 4x5}}\\
\cmidrule(lr){2-5}\cmidrule(lr){6-9}\cmidrule(lr){10-13}\cmidrule(lr){14-17}
& RAcc & CAcc & R-ARI & C-ARI
& RAcc & CAcc & R-ARI & C-ARI
& RAcc & CAcc & R-ARI & C-ARI
& RAcc & CAcc & R-ARI & C-ARI\\
\midrule
SCC
& \best{86} & \best{65} & 0.64 & 0.32
& 62 & \second{51} & 0.28 & 0.00
& \second{82} & \best{73} & 0.56 & 0.19
& \best{81} & 50 & 0.56 & 0.20\\
SCCRP
& \second{82} & \second{63} & 0.58 & 0.30
& \second{62} & \best{55} & 0.28 & 0.12
& \second{82} & \second{72} & 0.54 & 0.17
& \second{78} & \best{53} & 0.50 & 0.19\\
SCCP-RS
& 77 & 57 & 0.41 & 0.26
& 62 & 49 & 0.22 & 0.00
& 80 & 68 & 0.49 & 0.10
& 73 & 44 & 0.38 & 0.07\\
SCCP
& 82 & 57 & 0.53 & 0.26
& \best{64} & 49 & 0.28 & 0.00
& \best{84} & 68 & 0.56 & 0.20
& 76 & \second{45} & 0.43 & 0.20\\
\bottomrule
\end{tabular}
\end{table*}

\section{Results and Discussion}

\subsection{Newsgroups Results}

Table~\ref{clustering_performance} reports the performance of all methods across four word-document matrices derived from the Newsgroups dataset. These results show that SCCRP occasionally sacrifices a small amount of accuracy, but it achieves best or second-best results in several cases, indicating that it is a strong and efficient alternative to the standard method. In contrast, SCCP-RS has the weakest clustering performance and does not offer any runtime benefits over SCCP. This is likely due to high sparsity of the dataset, which limits the effectiveness of random sampling. Column clustering performance was lower than row clustering across all methods, consistent with the inherent difficulty of word grouping.

\begin{figure}[t]
\centering
\vspace{8pt}
\begin{minipage}[b]{0.48\linewidth}
\centering
\includegraphics[width=\linewidth]{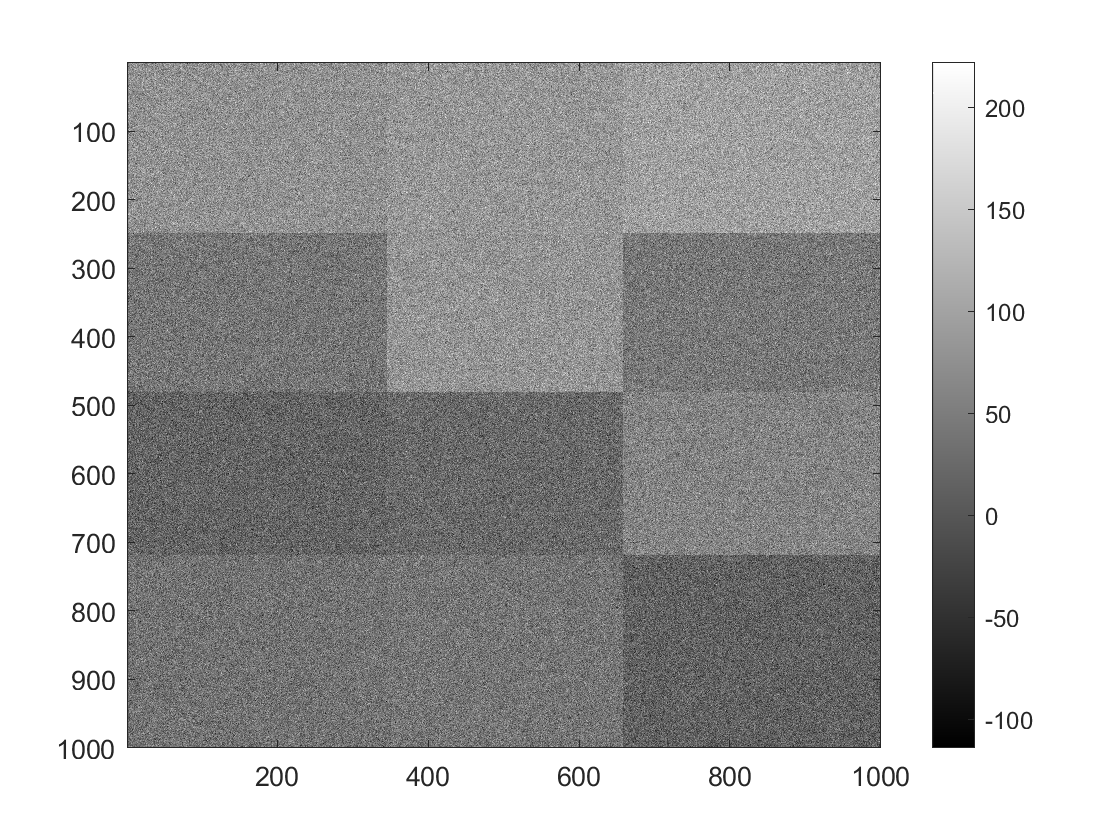}\\[-1mm]
\small a) Original matrix
\end{minipage}
\hfill
\begin{minipage}[b]{0.48\linewidth}
\centering
\includegraphics[width=\linewidth]{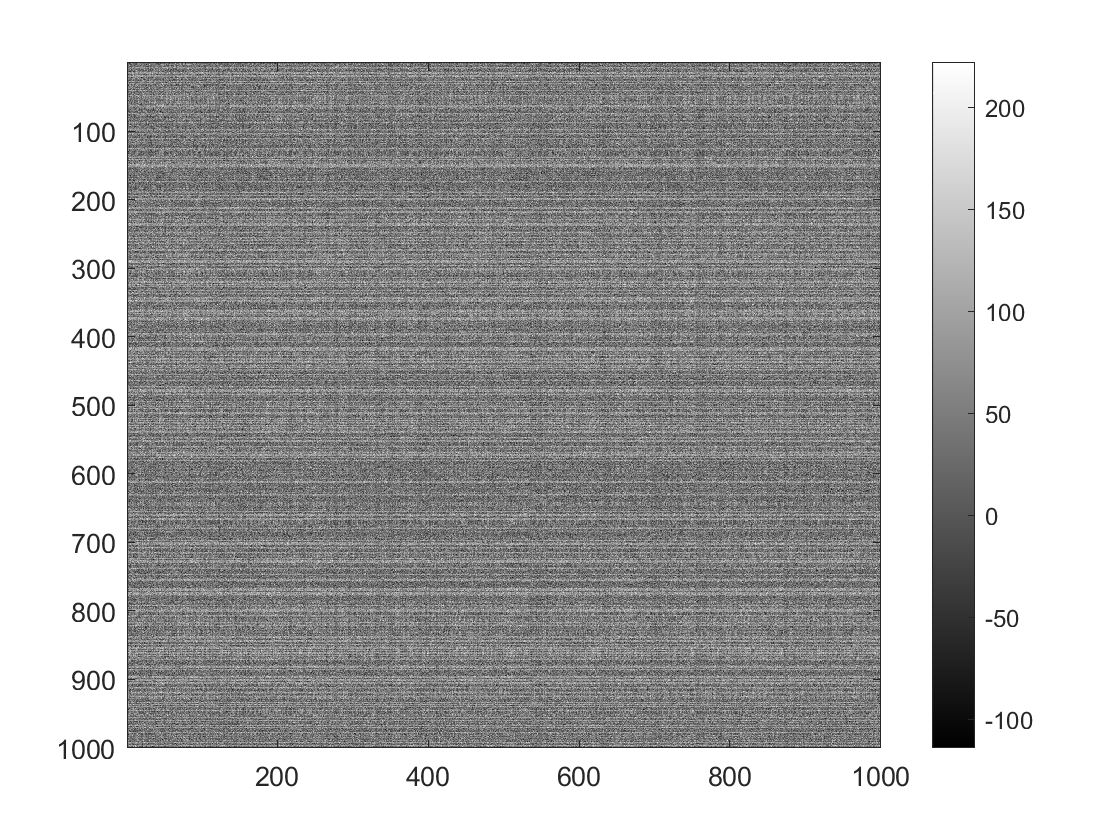}\\[-1mm]
\small b) Shuffled to hide structure
\end{minipage}
\vspace{1.2ex}

\begin{minipage}[b]{0.48\linewidth}
\centering
\includegraphics[width=\linewidth]{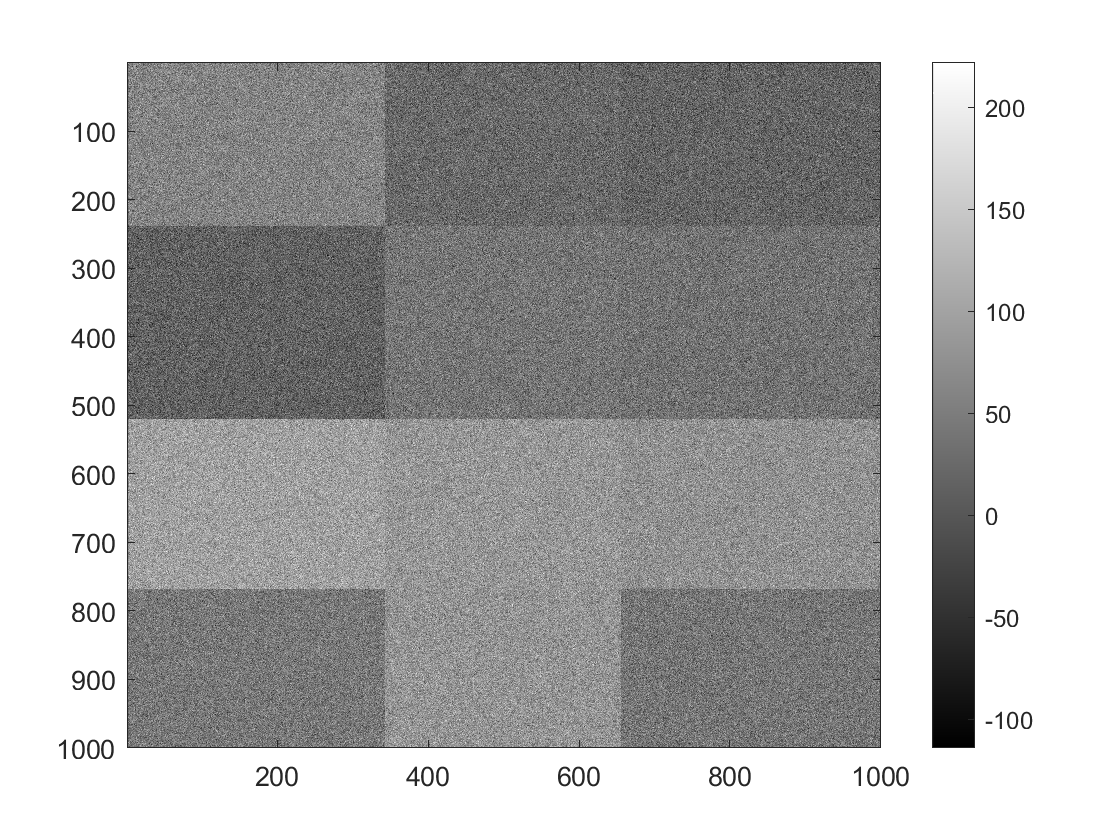}\\[-1mm]
\small c) SCCRP clustering result
\end{minipage}
\hfill
\begin{minipage}[b]{0.48\linewidth}
\centering
\includegraphics[width=\linewidth]{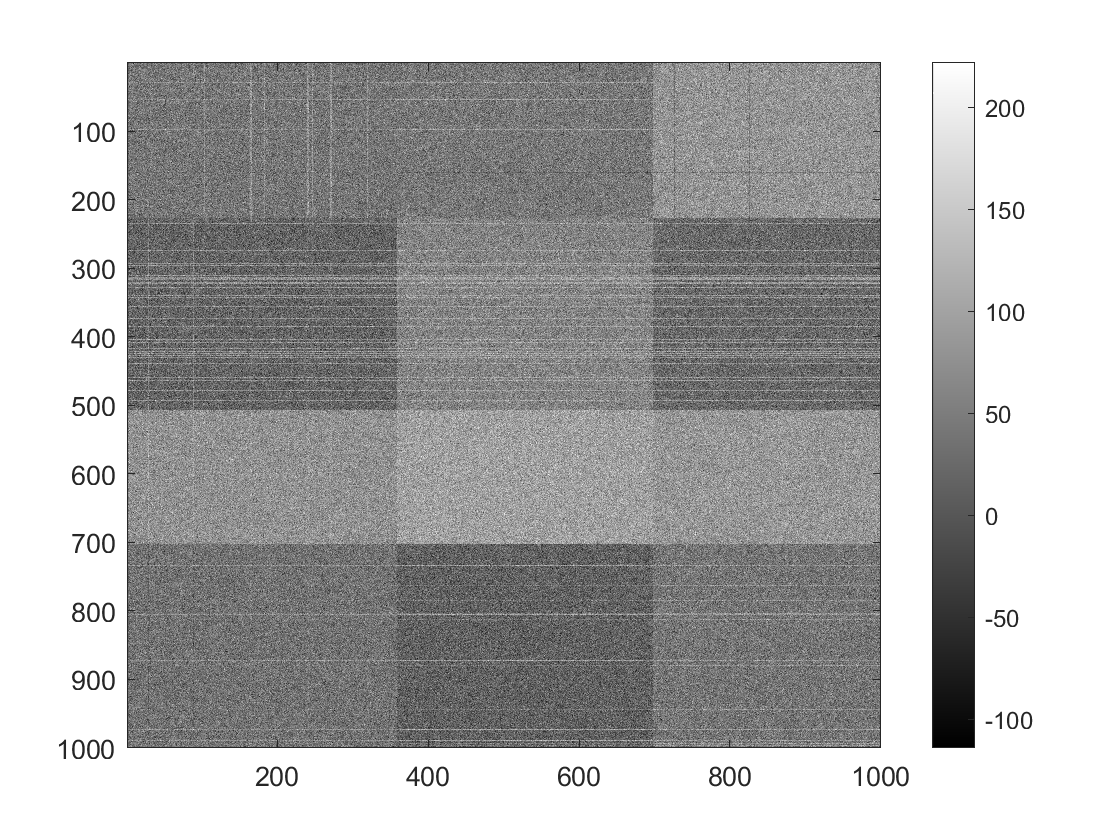}\\[-1mm]
\small d) SCCP-RS clustering result
\end{minipage}
\vspace{16pt}
\caption{Cluster patterns on a $1k\times1k$ synthetic checkerboard dataset (a), after shuffling (b), applying SCCRP (c), SCCP-RS with $p=0.4$ (d). Sorted based on the row and column labels assigned by SCCP-RS and SCCRP, and plotted the result. Each plot shows grayscale intensity of matrix entries.}
\label{cluster_patterns}
\end{figure}

As presented in Table~\ref{runtime_newsgroup}, the original SCC algorithm is consistently the slowest due to its reliance on full SVD. SCCRP and SCCP significantly reduce runtime while maintaining comparable performance. SCCP has the fastest runtime on the largest dataset (0.9 sec compared to SCC's 24.2 sec; a $27\times$ speedup). SCCP-RS does not offer clear runtime advantages and remains very similar to SCCP. These results demonstrate the importance of choosing suitable co-clustering strategies for the data structure. Figure~\ref{newsgroup_runtime} shows runtime
trends.

\subsection{Synthetic Datasets Results}

Figure~\ref{cluster_patterns} presents co-clustering results on synthetic datasets constructed with known checkerboard bicluster structure. Both SCCRP and SCCP-RS successfully recover the underlying patterns. Although SCCP-RS demonstrates slightly lower accuracy compared to SCCRP, it achieves a substantial reduction in runtime. This trade-off is illustrated in Figure~\ref{synthetic_runtime}. These results highlight the significant runtime advantage of the randomized methods over the baseline SCC. While SCCRP delivers clustering performance that is more comparable to SCC, making it a favorable middle ground between accuracy and efficiency, SCCP-RS is well-suited for scenarios where computational efficiency is critical.

\begin{figure}[t]
\centering
\begin{subfigure}[t]{0.48\linewidth}
\centering
\includegraphics[width=\linewidth]{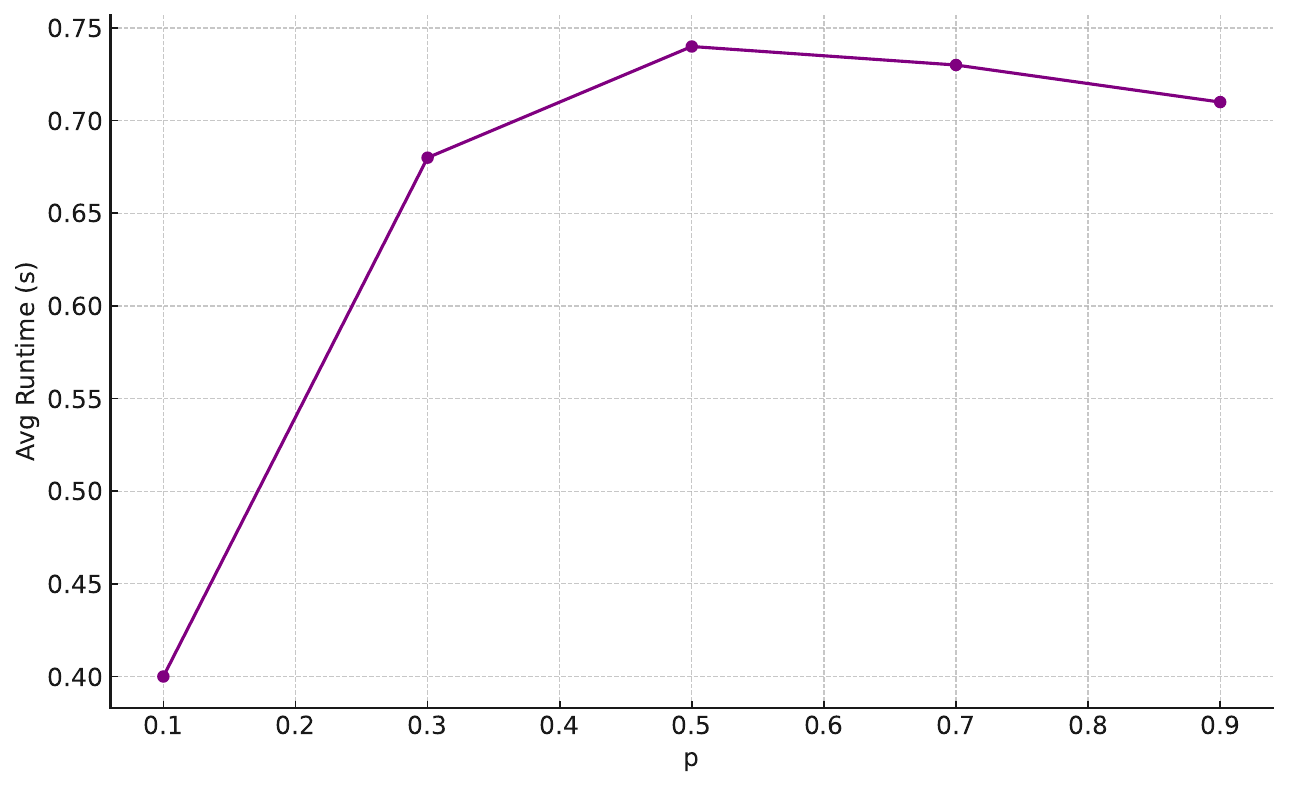}
\caption{NG: Runtime vs $p$}
\end{subfigure}
\hfill
\begin{subfigure}[t]{0.48\linewidth}
\centering
\includegraphics[width=\linewidth]{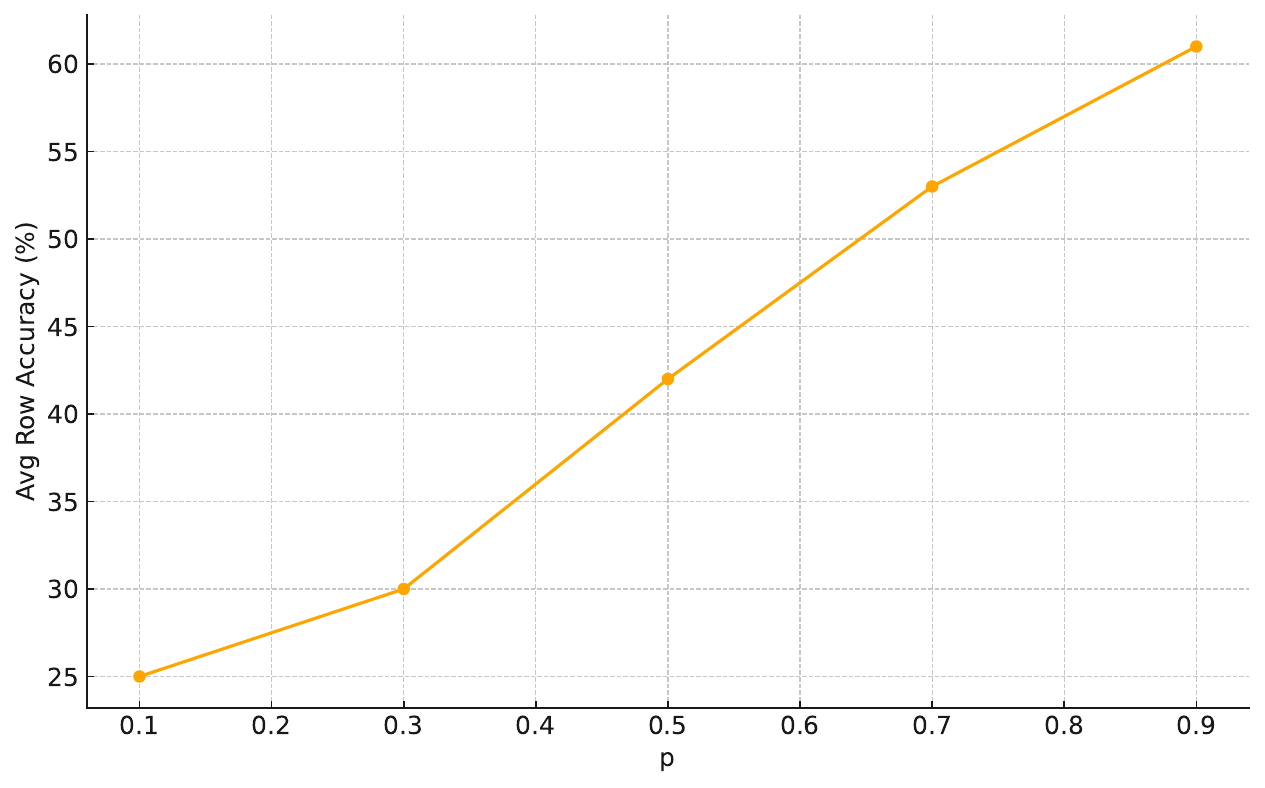}
\caption{NG: Accuracy vs $p$}
\end{subfigure}
\begin{subfigure}[t]{0.48\linewidth}
\centering
\includegraphics[width=\linewidth]{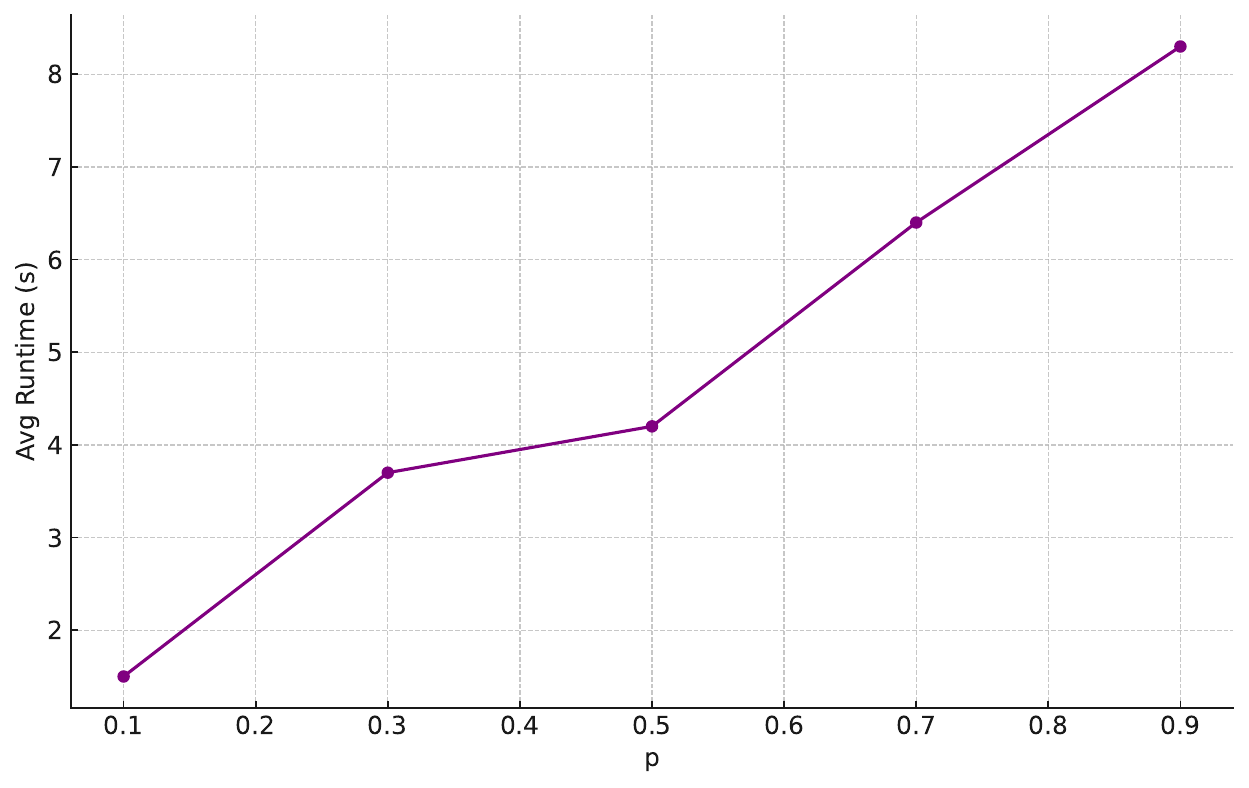}
\caption{CB: Runtime vs $p$}
\end{subfigure}
\hfill
\begin{subfigure}[t]{0.48\linewidth}
\centering
\includegraphics[width=\linewidth]{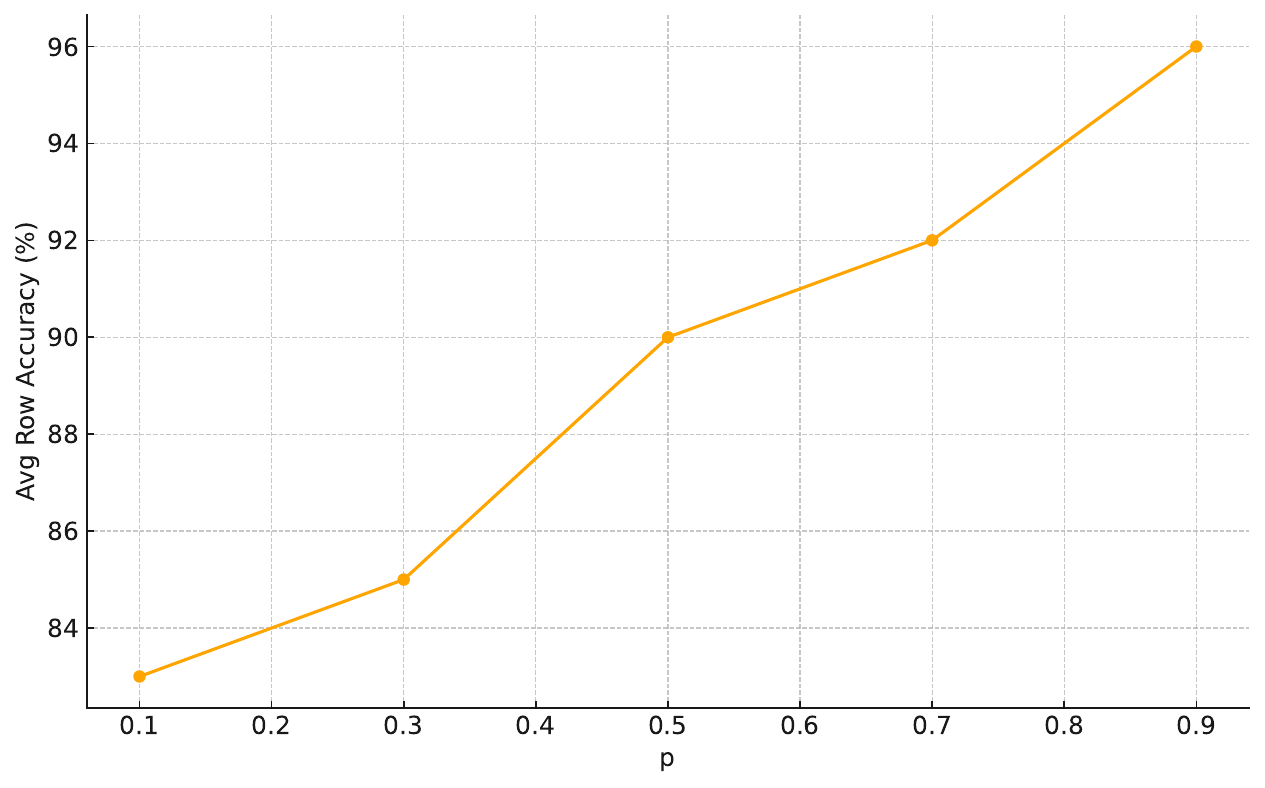}
\caption{CB: Accuracy vs $p$}
\end{subfigure}
\caption{Effect of the random sampling ratio $p$ on runtime and clustering accuracy for the sparse Newsgroups (NG) and dense Checkerboard (CB) datasets, which are comparable in size and have the same numbers of row and column clusters.}
\label{p_effect}
\end{figure}

\subsection{Random Sampling Effect}

Figure~\ref{p_effect} illustrates the effect of random sampling probability $p$ on efficiency and accuracy of SCCP-RS compared on both datasets. For the dense checkerboard dataset, decreasing $p$ from 1 to 0.1 leads to a reduction in runtime, demonstrating the computational benefits of random sampling. However, this efficiency gain comes at the cost of reduced clustering accuracy. In contrast, the sparse Newsgroups dataset shows a less consistent runtime pattern, which suggests that when the data is already sparse, moderate levels of random sampling may not reduce runtime further. However, accuracy still improves steadily with larger $p$. These observations emphasize that the choice of $p$ should be guided by the structure of data and the specific trade-off preferences in the target application.

\subsection{Estimated Scalability on Larger Matrices}

To better understand the behavior on larger datasets, we estimate runtime on larger matrices by extrapolating from the largest Newsgroups matrix, $3843\times4158$. This analysis provides an approximate indication of how the methods are expected to behave under larger input sizes. For the full-SVD baseline, we use the standard dense SVD scaling $O(nm\min(n,m))$. For SCCRP, SCCP-RS, and SCCP, we use the observed near-linear dependence on the number of matrix entries under fixed target rank and number of power iterations, and comparable sparsity. Table~\ref{estimates} reports these estimates.

\begin{table}[t]
\centering
\caption{Estimated runtime in seconds for larger square matrices, extrapolated from the largest Newsgroups runtime result. These values are projected scalability estimates.}
\label{estimates}
\scriptsize
\begin{tabular}{ccccc}
\toprule
\textbf{Matrix size} & \textbf{SCC} & \textbf{SCCRP} & \textbf{SCCP-RS} & \textbf{SCCP}\\
\midrule
$5000\times5000$ & 49.3 & 8.0 & 1.7 & 1.4\\
$10000\times10000$ & 394.1 & 31.9 & 6.8 & 5.6\\
$20000\times20000$ & 3152.7 & 127.7 & 27.0 & 22.5\\
\bottomrule
\end{tabular}
\end{table}

\subsection{Limitations}

Our experiments use moderate-size 20 Newsgroups matrices rather than production-scale corpora, so they do not establish largescale performance. Also it does not provide ground-truth word categories, and evaluation is indirect measure of column-clustering quality. The randomized SVD and sampling steps use existing randomized linear-algebra tools; our contribution is integration and evaluation within nonsquare normalized spectral co-clustering. Finally, clustering quality depends on normalization, eigengap structure, $k$-means initialization, and row-column signal strength, so our findings remain empirical.

\bibliographystyle{IEEEtran}
\bibliography{references}

\end{document}